\documentclass{article}

\usepackage{PRIMEarxiv}

\usepackage[utf8]{inputenc} 
\usepackage[T1]{fontenc}    
\usepackage{hyperref}       
\usepackage{url}            
\usepackage{booktabs}       
\usepackage{amsfonts}       
\usepackage{nicefrac}       
\usepackage{microtype}      
\usepackage{lipsum}
\usepackage{fancyhdr}       
\usepackage{graphicx}       
\graphicspath{{media/}}     
\usepackage{bbm}
\usepackage{amsmath}
\usepackage{amssymb}
\usepackage{multirow}
\usepackage{appendix}

\usepackage[ruled,vlined]{algorithm2e}
\usepackage{xcolor}

\title{Self-Supervised Representation Learning With MUlti-Segmental Informational Coding (MUSIC)
}

\author{
  Chuang Niu and Ge Wang \\
  Rensselaer Polytechnic Institute \\
  Troy, New York, USA \\
  \texttt{\{niuc, wangg6\}@rpi.edu} \\
}

\begin{document}
\maketitle

\begin{abstract}
Self-supervised representation learning maps high-dimensional data into a meaningful embedding space, where samples of similar semantic contents are close to each other.
Most of the recent representation learning methods maximize cosine similarity or minimize the distance between the embedding features of different views from the same sample usually on the $l2$ normalized unit hypersphere.
To prevent the trivial solutions that all samples have the same embedding feature, various techniques have been developed, such as contrastive learning, stop gradient, variance and covariance regularization, etc.
In this study, we propose MUlti-Segmental Informational Coding (MUSIC) for self-supervised representation learning.
MUSIC divides the embedding feature into multiple segments that discriminatively partition samples into different semantic clusters and different segments focus on different partition principles.
Information theory measurements are directly used to optimize MUSIC and theoretically guarantee trivial solutions are avoided.
MUSIC does not depend on commonly used techniques, such as memory bank or large batches, asymmetry networks, gradient stopping, momentum weight updating, etc, making the training framework flexible.
Our experiments demonstrate that MUSIC achieves better results than most related Barlow Twins and VICReg methods on ImageNet classification with linear probing, and requires neither deep projectors nor large feature dimensions. Code will be made available.
\end{abstract}


\section{Introduction}

Self-supervised representation learning (SSRL) has been a core task in machine learning \cite{bengio2013representation} and seen rapid grogress over past years.
Deep neural networks pre-trained on large-scale unlabeled datasets via SSRL have demonstrated desirable characteristics, such as strong robustness \cite{hendrycks2019using} and generalizablity \cite{mohseni2020self}, and improving various down-stream tasks especially when annotations are scarce.
An effective approach for SSRL is to enforce semantically similar samples (i.e., different transformations from the same instance) close to each other in the embedding space.
Simply maximizing the similarity or minimizing the Euclidean distance between embedding features of similar semantic samples tends to produce trivial solutions, e.g., all samples have the same embedding.
Recently, various excellent methods have been proposed to learn meaningful representations features and avoid trivial solutions.
Contrastive learning~\cite{hadsell2006dimensionality, van2018representation} based methods, such as SimCLR \cite{simclr} and MoCo \cite{moco}, have achieved great successes by additionally minimizing the similarity between embeddings of reference and negative samples, which requires either relatively large batches or a memory bank \cite{wu2018unsupervised, misra2020self} of negative samples during training.
To avoid using negative samples during training, BYOL \cite{grill2020bootstrap} and SimSiam \cite{Chen_2021_CVPR} introduce some clever techniques, such as asymmetry network architecture with additional predictor head and stop gradients, etc.
Subsequent theoretical analysis \cite{understand, zhang2021does, richemond2020byol, tian2021understanding} have demonstrated why these techniques can avoid trivial solutions and learn meaningful representations from different aspects.
Clustering-based methods DeepCluster \cite{caron2018deep}, SELA \cite{asano2019self}, SwAV \cite{caron2020unsupervised} alternatively compute the cluster assignment of one view and optimize the network to predict the same assignment from other views of the same sample, where trivial solutions can be avoided via even assignment of samples over different clusters in a non-differentiable manner.
In another direction, W-MSE \cite{ermolov2021whitening} and Barlow~Twins \cite{zbontar2021barlow} propose to drive self- or cross-correlation matrices towards the identity matrix, learning meaningful features without requiring the asymmetry design or a large batch of negative samples.
Along the same direction, VICReg \cite{bardes2021vicreg} constructs a loss function with three terms, i.e., invariance, variance, and covariance constraints that can explicitly avoid trivial solutions.

\begin{figure*}[bt!]
    \centering
    \includegraphics[width=1\textwidth]{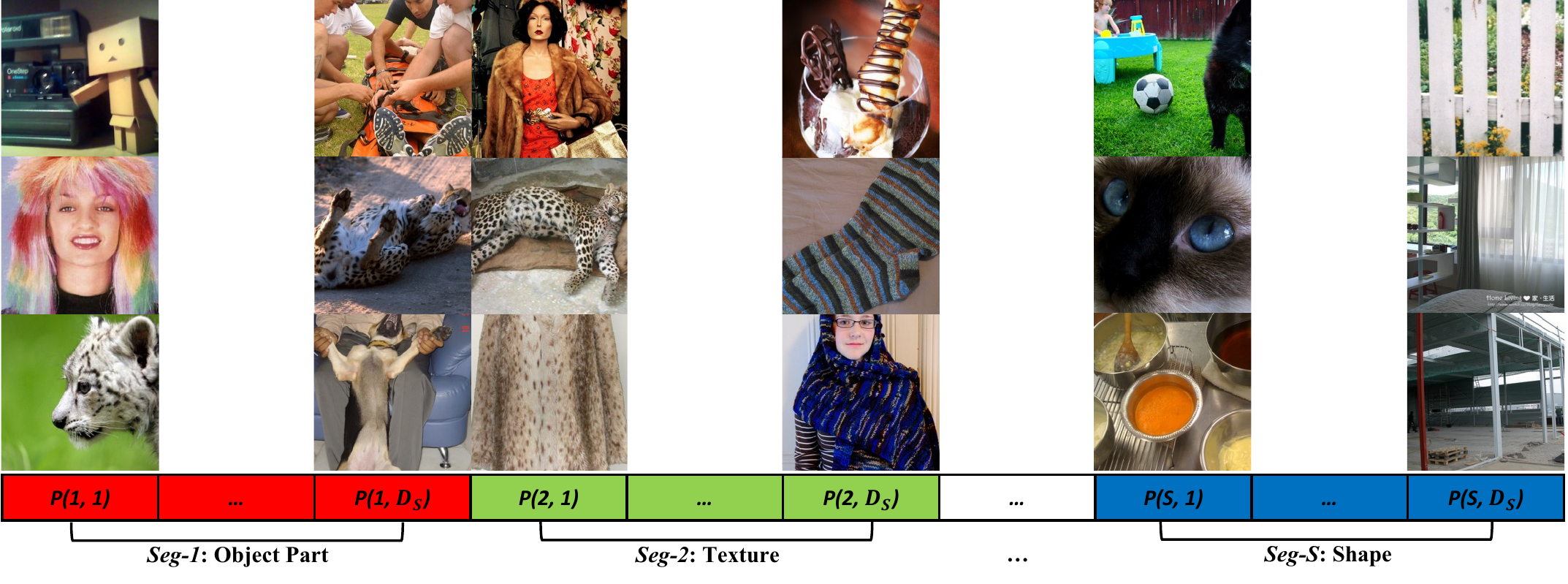}
    \caption{Partition of an MUSIC feature vector. An image should be represented by multiple attributes, such as general object parts, textures, shapes, etc. Motivated by this observation, MUSIC divides the embedding feature vector into multiple segments (Seg-1, Seg-2, ..., Seg-S); for example, here we show three different segments colored in red, green, and blue colors respectively to represent different attributes. The general attribute of each segment consists of multiple instantiations, and different instantiated attributes within the same segments are discriminative from each other. For example, Seg-2 represents texture, each unit in Seg-2 represents a specific texture, like dot texture, stripe texture, etc. Here three samples are shown over each unit. The value $p(s, d)$ in each unit denotes the probability of an image belongs to the $d^{th}$ instantiated attribute of $s^{th}$ segment, see Methodology section for more details.}
    \label{fig_vector}
\end{figure*}

Fundamentally different from the current SSRL methods that normalize embedding features onto the unit hpyersphere via l2 norm and use cosine similarity as the measurements,
we propose a new coding scheme, named MUlti-Segmental Informational Coding (MUSIC).
The motivation is based on the observation that an object can be represented in multiple attributes, such as object parts, textures, shapes, etc.
As shown in Fig. \ref{fig_vector}, the embedding vector is divided into a number of $S$ segments (i.e., Seg-1, Seg-2, ..., Seg-S), different segments represent different attributes; e.g., Seg-1, Seg-2, and Seg-3 represent object part, texture, and shape, respectively. Each segment instantiates a number of $D_S$ different features; e.g., Seg-2 represents samples in different textures (dot texture, stripe texture, etc.), see Fig. \ref{fig_vector} for more descriptions.
This means that different instantiated features within each segment are discriminative from each other.
Then, a specific instance can be uniquely represented by a set of well defined attributes. 
To automatically learn such MUSIC embeddings from unlabeled datasets, we introduce an entropy-based loss function.
Furthermore, we  theoretically analyze based on information theory why  meaningful features can be learned while trivial solutions are avoided.


The characteristics of MUSIC are as follows.
(1) MUSIC allows an information theory based representation learning framework in a novel way.
Theoretical analysis ensures that the optimized MUSIC embedding features are transform-invariance, discriminative, diverse, and non-trivial.
(2) Similar to Barlow Twins and VICReg, the presented MUSIC method does not require the asymmetry network architecture with an extra predictor module, a large batch size of contrastive samples, a memory bank, gradient stopping, or momentum updating.
(3) Different from existing methods, empirical results show that MUSIC does not depend on very high dimension of embedding features or a deep projection head, efficiently reducing the memory and computation cost.
(3) Extensive results show that MUSIC achieves better results than the state-of-the-art Barlow Twins and VICReg methods under the same conditions in terms of linear probing on the common ImageNet dataset.

\section{Methodology}
\label{sec_method}

\subsection{Architecture}

\begin{figure*}[h]
    \centering
    \includegraphics[width=1\textwidth]{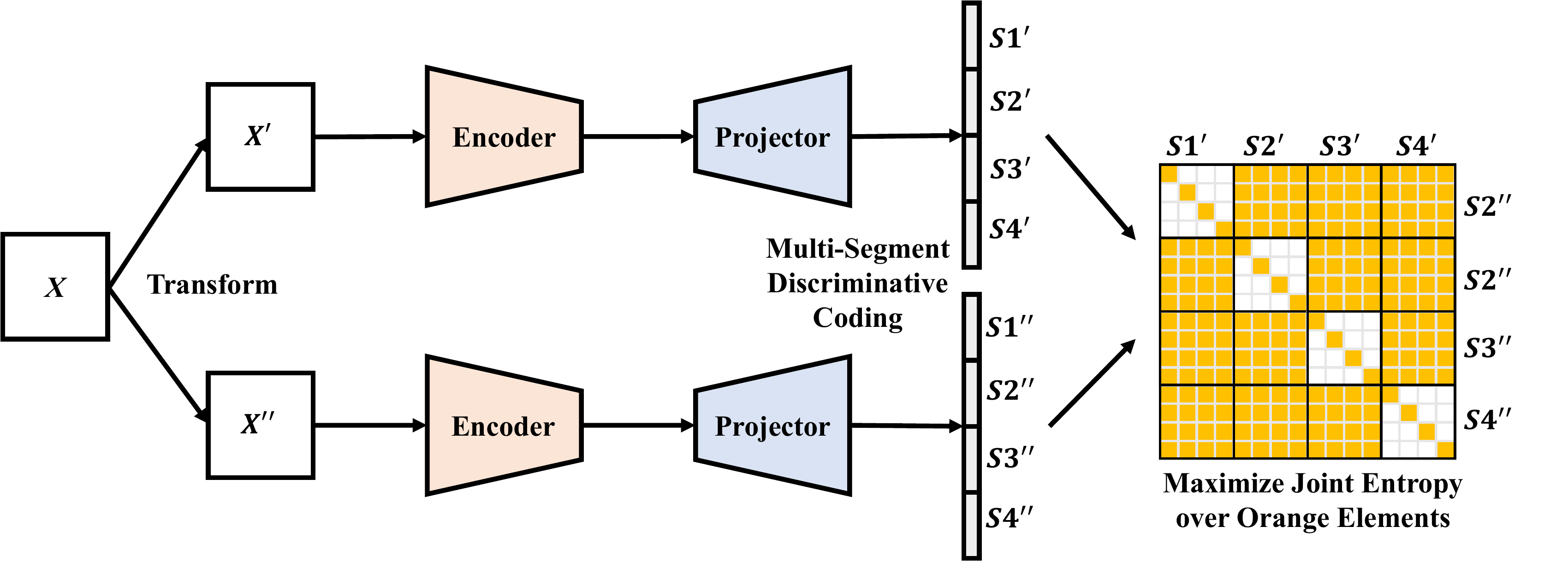}
    \caption{SSRL framework through multi-segmental informational coding optimized with maximum entropy.}
    \label{fig_ssl}
\end{figure*}

Similar to W-MSE and Barlow Twins, in this study we adopt a twin architecture that uses the same network for two branches without requiring any asymmetric design, as shown in Fig. \ref{fig_ssl}.
During training, input images $X=\{x_i\}_{i=1}^N$ are mapped to two distorted sets $X'=\{x'_i\}_{i=1}^N$ and $X''=\{x''_i\}_{i=1}^N$, where $N$ is the batch size.
The common transformation distribution, i.e., random crops combined with color distortions, the same as that in \cite{bardes2021vicreg}, is used to generate training samples .
Then, the two batches of distorted images $X'$ and $X''$ are respectively fed to two branches, each of which consists of an encoder $F(\cdot; \theta_F)$ and a projector $P(\cdot; \theta_P)$.
The output of the encoder is commonly used as the representation feature. The projection head maps the representation feature into the embedding sapce during training.
Note that the presented method is not limited to this twin architecture, which can be extended to two branches with different parameters or of heterogeneous networks, or even different input modalities (e.g., text, audio, etc.).


\subsection{Multi-Segmental Informational Coding}

Here we introduce the MUlti-Segmental Informational Coding (MUSIC) for self-supervised representation learning.
The embedding features of the two branches are denoted as: $z'_i = P(F(x'_i; \theta_F); \theta_P) \in R^D$, and $z''_i = P(F(x''_i; \theta_F); \theta_P) \in R^D$, where $D$ is the feature dimension.
Most of the existing SSL methods normalize embedding features in the L2 norm and then maximize the cosine similarity between the normalized results.
Motivated by the observation described in Fig. \ref{fig_vector}, we divide the embedding feature $z_i$ into multiple segments, denoted by $z_i(s, d), s=1,\cdots, S, d=1,\cdots, D_S$, where $S$ is the number of segments, $D_S$ is the dimension of each segment, and $D = D_S \times S$ gives the dimension of the whole embedding space. Although the current version of MUSIC scheme evenly splits the embedding vector, it could be in principle extended to uneven configurations.
Each segment is normalized to a probability distribution $p'_i(s', d')$ over $D_S$ instantiated attributes using the softmax function, i.e.,
\begin{equation}
    p'_i(s', d') = \frac{\exp(z'_i(s, d))}{\sum_{k=1}^{D_S} \exp(z'_i(s, k))},
    \label{eq_prob}
\end{equation}
The probability distribution $p''_i(s, d)$ for the other branch is computed in the same way.
Thus, the MUSIC scheme can be interpreted as a combination of multiple classifiers or cluster operators that implement different classification criteria learned in a data-driven fashion.

\subsection{Entropy Loss}
Based on the probability distributions over multiple segments, we first compute the empirical joint distribution $p(s', s'', d', d'')$ between the embedding features of two transformations over a batch of samples (similar to \cite{ji2019invariant}):
\begin{equation}
    p(s', s'', d', d'') = \frac{1}{N}\sum_{i=1}^N p'_i(s', d') p''_i(s'', d''),
    \label{eq_joint}
\end{equation}
With the empirical joint distribution, two versions of the loss function can be defined.
We denote the first version $L_{ent}$ as a pure entropy-based loss function:
\begin{equation}
    L_{ent} = \frac{1}{S^2} \sum_{s'=1}^{S}\sum_{s''=1}^S\sum_{d'=1}^{D_S}\sum_{s''=1}^{D_S} (1 - \mathbbm{1}(s'=s'', d'\ne d'')) p(s', s'', d', d'') \log(p(s', s'', d', d'')),
\end{equation}
where $\mathbbm{1}(s'=s'', d'\ne d'')$ is an indicator function that equals to 1 if $s' = s''$ and $d' \ne d''$, otherwise it is equal to 0. The empirical joint distribution can be denoted by a block matrix in Fig. \ref{fig_ssl}, where $(1 - \mathbbm{1}(s'=s'', d'\ne d''))$ means only keeping diagonal elements and elements of the off-diagonal blocks, as indicated by the orange area. Therefore, minimizing this loss function maximizes the joint entropy over the selected elements.
In the next subsection, we show that this single loss function allows to learn meaningful features.

To enhance the transformation invariance of features, we introduce an additional term to maximize the inner product between the embedding features from two transformations. Then, the second version of the loss function is defined as
\begin{equation}
    \begin{split}
        L = L_{ent} - \lambda \frac{1}{NS} \sum_{i=1}^{N}\sum_{s=1}^S\sum_{d=1}^{D_S} \log(p'_i(s, d) p''_i(s, d)),
    \end{split}
\end{equation}
where $\lambda$ is a balancing factor. By default we set $\lambda = 1$ and found that $\lambda$ is not required to be a very small or large number for balancing.
Since $p'_i(s, d)$ and $p''_i(s, d)$ are the probabilities, maximizing their inner product imposes the network to make consistent assignment over all segments between two transformations of the same image. As a result, each segment is encouraged to be a one-hot vector for the maximum inner product.
Clearly, this additional term promotes transformal invariance and confident assignments over different attributes.
One difference of this term from the entropy loss term is the sample-specific constraint while entropy is a statistical measure.

Our proposed method can be easily implemented, with a PyTorch-style pseudo code in the Appendix.
In the following subsection, let us analyze why the entropy loss optimizes meaningful embedding features as illustrated in Fig. \ref{fig_vector}.

\subsection{Analysis}
\label{sec_analysis}

The entropy loss function consists of two parts, including the entropy over diagonal elements and the entropy over the elements of off-diagonal blocks illustrated in Fig. \ref{fig_ssl}, formally denoted as
\begin{equation}
\begin{split}
        L_{ent} = &\frac{1}{S}\sum_{s',s'', s'=s''}\sum_{d', d'', d'=d''} p(s', s'', d', d'') \log(p(s', s'', d', d'')) \\
            & + \frac{1}{S(S-1)}\sum_{s',s'', s'\ne s''}\sum_{d', d''} p(s', s'', d', d'') \log(p(s', s'', d', d'')).
\end{split}
\end{equation}
For the first part, it can be demonstrated that its optimal solution is $\forall s, d, p'_i(s, d) = p''_i(s, d)$, $p'_i(s, d)$ and $p''(s, d)$ are one-hot vectors, and $\frac{1}{N}\sum_{d=1}^{N} p_i(s, d) = \frac{1}{D_S}$. The prof can be found in the Appendix.
For the second part, it is intuitive that the optimal solution to maximize the joint entropy over the off-diagonal block items is $\forall s', s'', d', d'', s' \ne s'', p(s', s'', d', d'') = \frac{1}{({D_S})^2}$; i.e., a batch of samples are evenly assigned over each off-diagonal block.

\textbf{Transform Invariance}:
The solution that $p''(s, :)$ and $p''(s, :)$ are one-hot vectors and equal to each other means that the learned MUSIC embeddings are invariant to transformations, and a sample tends to be confidently represented by a single instantiated attribute within each and every segment.

\textbf{Non-trivial Solution}:
The solution that $\frac{1}{N}\sum_{d=1}^{N} p_i(s, d) = \frac{1}{D_S}$ means that each segment evenly partition a batch of samples over $D_S$ instantiated attributes. Since $p'_i(s, d)$ and $p''(s, d)$ are one-hot vectors, the trivial solution that all samples have the same embedding features or assigned to the same attribute for each segment can be avoided.

\textbf{Minimum Redundancy}:
As described in Fig. \ref{fig_vector}, different segments of the MUSIC embedding vector are expected to focus on complementary attributes.
In other words, the redundancy or mutual information between any two segments should be minimized, which is a popular measure for feature selection \cite{peng2005feature}.
Here we can demonstrate the redundancy or mutual information between any two segments is minimized when the optimal solution is obtained.
Specifically, the mutual information $I(s', s'')$ between any segments $s'$ and $s''$ is
\begin{equation}
    \begin{split}
        I(s', s'') =& H(s') + H(s'') - H(s', s'') \\
                   =& -\sum_{d'=1}^{D_S} p'(s', d') log(p'(s', d')) - \sum_{d''=1}^{D_S} p''(s'', d'') log(p''(s'', d'')) \\
                    &+ \sum_{d'=1}^{D_S}\sum_{d''=1}^{D_S} p(s', s'', d', d'') log(p(s', s'', d', d'')) \\
                   =& -log\frac{1}{D_S} - log\frac{1}{D_S} + log\frac{1}{(D_S)^2} = 0.
    \end{split}
\end{equation}
Given that the features within each segment are naturally exclusive from each other, MUSIC embedding features are both discriminative and diverse.
The redundancy constraint has been studied for W-MSE \cite{ermolov2021whitening}, Barlow Twins \cite{zbontar2021barlow}, and VICReg \cite{bardes2021vicreg} by minimizing the covariance in a linear manner. In contrast, our entropy-based loss function reduces the redundancy in a non-linear way.
Also, it can be derived that the optimal MUSIC embedding features have zero covariance between any two features in different segments and negative covariance between the features within the same segment. More details can be found in the Appendix. 


\textbf{Contrastive Learning}:
Contrastive learning has proven very effective for representation learning by maximizing the similarity between different transformations of the same instance and minimizing the similarity between the reference and other instances.
Here it can be seen that MUSIC is consistent to contrastive learning in a novel way.
Specifically, the optimal MUSIC embedding can totally encode $(D_S)^S$ different samples.
In our default settings $D_S=80, S=102$ (See the Empirical Analysis below for more details), MUSIC can represent $80^{102}$ different samples.
Maximizing the joint entropy is to evenly assign a batch of samples into all possible embeddings, which means that the embedding features of all instances are enforced to be different from each other like in contrastive learning, given the sufficiently large coding capacity.
Therefore, the difference is that contrastive learning differentiates instances by directly enforcing their features to be dissimilar, while MUSIC statistically assigns instances with different assignment codes.

In a word, the MUSIC embedding feature optimized with the entropy-based loss is transform-invariant, non-trivial, dicriminative, and diverse. 

\section{Implementation Details}
For fair comparison, we followed the same settings in VICReg \cite{bardes2021vicreg}.
Specifically, the standard ResNet-50 backbone \cite{He_2016_CVPR} was used as the encoder that outputs a representation vector of 2,048 units. We used the same training settings including the data augmentation (random cropping, horizontal flip, color jittering, grayscale, Gaussian blur, solarization, with the same parameters in \cite{bardes2021vicreg}), the optimizer of LARS \cite{you2017large, goyal2017accurate} with a weight decay of $10^{-6}$ and the learning rate of $lr=batch\_size/256 \times base\_lr$, and the cosine decay schedule \cite{loshchilov2016sgdr} from 0 with 10 warmup epochs towards the final value of 0.002. Here we set the base learning rate $bast\_lr$ to 0.6. By default, we used a two-layer MLP projector (8,192-8,160), the number of segments $S=102$, the segment dimension $D_S=80$, and $D=D_S \times S=8,160$ (similar to the feature dimension used by VICReg and Barlow Twins).
The results  were respectively analyzed for different feature dimensions, depths of projectors, $batch\_size$, segment dimension $D_S$, and training epochs.
MUSIC introduces a single extra hyperparameter $D_S$, its effects on the performance was evaluated.
All experiments were conducted on the 1,000-classes ImageNet dataset, where labels were not used for self-supervised representation learning.

\section{Results}

\subsection{Linear and Semi-Supervised Evaluations on ImageNet}

\begin{table}[htp]
  \renewcommand{\arraystretch}{1.5}
  \renewcommand\tabcolsep{8pt}
 \caption{\textbf{Comparison of different methods on ImageNet linear classification}. Top-1 and Top-5 accuracies (in \%) of ResNet50 are reported.}
  \centering
  \begin{tabular}{c|cc}
     Methods         & Top-1 & Top-5  \\
    \hline
     Supervised           &   76.5 &  -  \\
     \hline
     MoCo         (2020)           &   60.6 &  -  \\
     PIRL         (2020)           &   63.6 & -  \\
     CPC v2       (2019)           &   63.8 & - \\
     CMC          (2019)           &   66.2 & - \\
     SimCLR       (2020)           &   69.3 & 89.0  \\
     MoCo v2      (2020)           &   71.1 & 90.1 \\
     SimSiam      (2020)           &   71.3 & - \\
     SwAV         (2020)           &   71.8 & - \\
     BYOL         (2020)           &   74.3 & 91.6 \\
     Barlow Twins (2021)           &   73.2 & 91.0 \\
     VICReg       (2022)           &   73.2 & 91.1 \\
     MUSIC         (Ours)           &  73.6 & 91.4 \\
    \hline
  \end{tabular}
  \label{tab:compare}
\end{table}

We followed the common evaluation protocol, i.e., linear probing that trains a linear classifier on top of the frozen representations, to evaluate the representations of self-supervised learning methods.
Being consistent with Barlow Twins \cite{zbontar2021barlow} and VICReg \cite{bardes2021vicreg}, a ResNet-50 backbone was trained with the batch size of 2,048 for 1,000 epochs on the training set of ImageNet, and the linear classification results including Top-1 and Top-5 accuracies of different methods on the evaluation set are reported in Table \ref{tab:compare}.
The difference from Barlow Twins and VICReg is that MIDC used a two-layer MLP projector (8,192-8,160) instead of three layers (8,192-8,192-8,192).
We followed exactly the same hyperparameters of VICReg \cite{bardes2021vicreg} for training the linear classifier.
The performance of VICReg is on par with another state of the art method BYOL that uses asymmetric techniques, such as an additional predictor and a momentum encoder.
The comparative results show that MUSIC achieves better results than Barlow Twins and VICReg, where all these three methods trained a twin architecture without using negative pairs or any asymmetric techniques.
Significantly relaxed constraints on the MUSIC architecture make it  adaptable to more applications like multi-modal mapping.
The different motivation behind and theoretical framework of MIDC lead to some unique characteristics that the projector depth, feature dimension, and batch size can be smaller than what used in the competing algorithms to obtain similar results.
More results are the in Empirical Analysis below.

\begin{table}[htp]
  \renewcommand{\arraystretch}{1.5}
  \renewcommand\tabcolsep{8pt}
 \caption{\textbf{Semi-Supervised Learning}. Top-1 and Top-5 accuracies (in \%) of classification on ImageNet. These results were obtained using ResNet50.}
  \centering
  \begin{tabular}{c|cccc}
     \multirow{2}{*}{Methods}      & \multicolumn{2}{c}{Top-1} & \multicolumn{2}{c}{Top-5} \\
                                   & 1\% & 10\% & 1\% & 10\% \\
    \hline
     Supervised                    &   25.4 &  56.4   &   48.4 &  80.4  \\
     \hline
     MoCo         (2020)           &   - &  -         &   57.2 &  83.8 \\
     SimCLR       (2020)           &   48.3 & 65.6    &   75.5 &  87.8 \\
     BYOL         (2020)           &   53.2 & 68.8    &   78.4 &  89.0 \\
     Barlow Twins (2021)           &   55.0 & 69.7    &   79.2 &  89.3 \\
     VICReg       (2022)           &   54.8 & 69.5    &   79.4 &  89.5 \\
     MUSIC         (Ours)           &   54.0 & 69.0    &   78.9 &  89.1 \\
    \hline
  \end{tabular}
  \label{tab:semi}
\end{table}

We also evaluated its effectiveness on semi-supervised learning.
Here the pretrained ResNet-50 with MUSIC was fine-tuned on subsets of ImageNet, including 1\% and 10\% of full ImageNet data respectively.
All the comparison methods used the same subset images.
Currently, MUSIC is not as good as Barlow Twins and VICReg in the semi-supervised learning settings, while it is better than BYOL and other compared methods.
Note that the current results of MUSIC were obtained by simply using the training parameters for Barlow Twins, while different methods usually used different hyperparamters to achieve their best results for this task. 
The compared methods did a grid search for different learning rates of the backbone and linear head and report the best results
In future, we plan to report more results optimized in a similar way.

\subsection{Empirical Analysis}

In this subsection, we evaluated the effects of different hyperparameters on the proposed MUSIC method and compared it with other SSRL methods. All methods were evaluated with linear classification on ImageNet.

\subsubsection{Effects of Epoch Number}

\begin{table}[htp]
  \renewcommand{\arraystretch}{1.5}
  \renewcommand\tabcolsep{8pt}
 \caption{\textbf{Comparison of different training epochs}. Top-1 accuracy (in \%) of linear results for linear classification on ImageNet were obtained using ResNet50.}
  \centering
  \begin{tabular}{c|cccccccc}
     Methods         & SimCLR & MoCo v2 & BYOL & SwAV & SimSiam & Barlow Twins & VICReg & MUSIC \\
    \hline
     100 epochs      &   66.5 & 67.4 & 66.5 &66.5 &  68.1 &  68.7 &  68.6   & 69.4  \\
     200 epochs      &   68.3 & 69.9 & 70.6 &69.1 &  70.0 &  - &  -  & 71.8 \\
     400 epochs      &   69.8 & 71.0 & 73.2 &70.7 &  70.8 &  - &  -  & 73.1  \\
     800 epochs      &   70.4 & 72.2 & 74.3 &71.8 &  71.3 &  - &  -  & 73.4  \\
    \hline
  \end{tabular}
  \label{tab:epochs}
\end{table}

The SSRL methods in different studies do not always use the same training epochs due to different computational environments.
Here MUSIC was evaluated on different training epochs as reported in Table \ref{tab:epochs}.
MUSIC is consistently better than most of existing methods on all different training epochs.
When the training epochs are small (100 and 200), MUSIC can converge to the best results.

\subsubsection{Effect of Batch Size}

\begin{table}[htp]
  \renewcommand{\arraystretch}{1.5}
  \renewcommand\tabcolsep{8pt}
 \caption{\textbf{Batch Size}. Top-1 accuracy (in \%) results for linear classification on ImageNet were obtained based on ResNet50 with 100 pretraining epochs.}
  \centering
  \begin{tabular}{c|ccc}
     Batch Size         & 512 & 1024 & 2048 \\
    \hline
     SimSiam             & 68.1 & 68.0 & 67.9 \\
     VICReg              & 68.2 & 68.3 & 68.6 \\
     MUSIC                & 68.3 & 69.3 & 69.4 \\
    \hline
  \end{tabular}
  \label{tab:batchsize}
\end{table}

\begin{table}[htp]
  \renewcommand{\arraystretch}{1.5}
  \renewcommand\tabcolsep{8pt}
 \caption{\textbf{Projector Depth}. Top-1 and Top-5 accuracies (in \%) of linear classification on ImageNet were obtained based on ResNet50 with 100 pretraining epochs.}
  \centering
  \begin{tabular}{c|ccc}
     Projector Depth         & 2 (8192-8160) & 3 (8192-8192-8160) & 4 (8192-8192-8192-8160) \\
    \hline
     Top-1               & 69.4 & 68.5 & 67.9 \\
     Top-5               & 89.3 & 88.3 & 87.9 \\
    \hline
  \end{tabular}
  \label{tab:projector}
\end{table}

\begin{table}[htp]
  \renewcommand{\arraystretch}{1.5}
  \renewcommand\tabcolsep{8pt}
 \caption{\textbf{Enhanced Transform Invariance Loss}. Top-1 and Top-5 accuracies (in \%) of linear classification on ImageNet were obtained based on ResNet50 with 100 pretraining epochs.}
  \centering
  \begin{tabular}{c|ccc}
     Loss         & Entropy & Entropy + Transform Invariance \\
    \hline
     Top-1               &  65.4 & 69.4 \\
     Top-5               &  86.9 & 89.3 \\
    \hline
  \end{tabular}
  \label{tab:loss}
\end{table}

\begin{table}[htp]
  \renewcommand{\arraystretch}{1.5}
  \renewcommand\tabcolsep{8pt}
 \caption{\textbf{Segment Dimension}. Top-1 and Top-5 accuracies (in \%) of linear classification on ImageNet were obtained based on ResNet50 with 100 pretraining epochs.}
  \centering
  \begin{tabular}{c|ccccc}
     $D_S$         & 32 & 64 & 80 & 96 & 128  \\
    \hline
     Top-1            & 67.8 & 69.1 & 69.4  &69.2 & 68.4 \\
     Top-5            & 88.5 & 89.1 & 89.3  &89.1 & 88.5 \\
    \hline
  \end{tabular}
  \label{tab:seg}
\end{table}

\begin{table}[h]
  \renewcommand{\arraystretch}{1.5}
  \renewcommand\tabcolsep{8pt}
 \caption{\textbf{Feature Dimension}. Top-1 accuracy (in \%) results for linear classification on ImageNet were obtained based on ResNet50 with 100 pretraining epochs.}
  \centering
  \begin{tabular}{c|ccccc}
     Feature Dimension         & 1024 (960) & 2048 (2000) & 4096 (4080) & 8192 (8160)  & 16384 (16320) \\
    \hline
     VICReg            &   62.4 & 65.1 & 67.3 &68.6    & 68.8  \\
     MUSIC              &   64.1 & 66.6 & 69.2 &69.4    & 69.1  \\
    \hline
  \end{tabular}
  \label{tab:dim}
\end{table}

Here we evaluated the performance of MUSIC on different batch sizes. The results in Table \ref{tab:batchsize} show that MUSIC is consistently better than the compared methods using different batch sizes.

\subsubsection{Effect of Projector Depth}

The existing methods \cite{zbontar2021barlow} require at least 3 layers of MLP as the projector for the best results.
However, MUSIC has a different behavior that a two-layer MLP achieves the best results as shown in Table \ref{tab:projector}.
These results may be due to the discriminability and diversity of MUSIC embeddings, making it easy to meaningfull representations.

\subsubsection{Effects of Loss Function}

As described in the Methodology section, optimizing the entropy loss only can avoid trivial solutions and learn meaningful representations. This theoretical analysis is consistent with the empirical results in Table \ref{tab:loss} that 65.4\% Top-1 was achieved using the entropy loss only, comparable to some methods reported in Table \ref{tab:compare}.
Adding the enhanced transform invariance term can significantly improve the performance, as also discussed in the Methodology section, the transform invariance can be further enhanced with this image-level constraint.

\subsubsection{Effect of Segment Dimension}

The effect of our unique hyperparameter, i.e., segment dimension, was underlined. Our experimental results of different segment dimensions in Table \ref{tab:seg} indicate that $D_s = 80$ achieved the best results, where the dimension of the whole embedding feature was kept the same. It can be seen that the performance is not sensitive to this hyperparameter.

\subsubsection{Effects of Feature Dimension}

In the previous studies for Barlow Twins and VICReg, it was found that increasing the feature dimension is very effective to improve the representation learning performance. It was also found that the feature dimension plays an important role in MUSIC. The results of different feature dimensions for VICReg and MUSIC are reported in Table \ref{tab:dim}. It can be seen that MUSIC achieves consistently better results than VICReg on different embedding feature dimensions. Importantly, when the embedding feature dimension is reasonably large (4,096 and 8,192), MUSIC achieves the best results and better than the best results of VICReg using the large dimension of 16,384. In practice, we found that the large embedding feature dimension (i.e., 16,384) significantly increases the computational and memory cost for Barlow Twins, VICReg, and MUSIC that compute the covariance or joint entropy matrix, which was also discussed in the Barlow Twins study \cite{zbontar2021barlow}. Therefore, MUSIC seems both efficient and effective.

\section{Conclusion}

We have presented the multi-segment informational coding (MUSIC) optimized with an entropy-based loss function for self-supervised representation learning. 
Experimental results show that MUSIC achieves equivalent or better representation learning results compared with the state of the art methods in terms of linear classification. The presented new framework ensures that MUSIC can avoid trivial solutions and learn discriminative and diverse features.
Interestingly, MUSIC has shown some unique characteristics that the projector can be a shallower MLP, the batch size and the embedding feature dimension can be smaller than that used in existing methods while achieving comparable or better results.
In the future, we will adapt and evaluate MUSIC to more downstream tasks, such as multi-modality tasks and medical applications.

\appendix

\makeatletter
\def\@seccntformat#1{\csname Pref@#1\endcsname \csname the#1\endcsname\quad}
\def\Pref@section{Appendix~}
\makeatother
\include{appendix_a}

\renewcommand\thefigure{\thesection.\arabic{figure}}    
\renewcommand{\theequation}{\thesection-\arabic{equation}}    
\definecolor{commentcolor}{RGB}{110,154,155}   
\newcommand{\PyComment}[1]{\ttfamily\textcolor{commentcolor}{\# #1}}  
\newcommand{\PyCode}[1]{\ttfamily\textcolor{black}{#1}} 

\section{PyTorch Pseudocode}
An example of PyTorch-style implementation for MUSIC is described in Algorithm \ref{algo:code}.

\begin{algorithm}[h]
\SetAlgoLined
    \PyComment{f: network function} \\
    \PyComment{lambda: weight on the transform invariance loss term} \\
    \PyComment{N: batch size} \\
    \PyComment{D: dimensionality of the embeddings} \\
    \PyComment{D\_S: dimensionality of each segment} \\
    \PyComment{S=D/D\_S: number of segments} \\
    \PyComment{} \\
    \PyComment{select: select the elements on diagonal and in off-diagonal blocks}\\
    \text{ } \\
    \PyCode{for i in loader:} \PyComment{load a batch with N samples} \\
    \Indp   
        \PyComment{two randomly augmented versions of x} \\
        \PyCode{x', x'' = augment(x)} \\
        \text{ } \\
        \PyComment{compute embeddings} \\
        \PyCode{z' = augment(x')} \\
        \PyCode{z'' = augment(x'')} \\
        \text{ } \\
        \PyComment{multi-segment discriminative coding} \\
        \PyCode{x' = torch.reshape(x', [N, -1, D\_S])} \PyComment{N$\times$S$\times$D\_S} \\
        \PyCode{x'' = torch.reshape(x'', [N, -1, D\_S])} \PyComment{N$\times$S$\times$D\_S} \\
        \PyCode{p' = torch.softmax(x', dim=2)} \PyComment{transform to probability} \\
        \PyCode{p'' = torch.softmax(x'', dim=2)} \PyComment{transform to probability} \\
        \text{ } \\
        \PyComment{compute transform invariance loss} \\
        \PyCode{loss\_TI = -torch.log((p'*p'').sum(dim=2)).mean()} \\
        \text{ } \\
        \PyComment{compute entropy loss} \\
        \PyCode{p' = torch.reshape(p', [N, D])} \PyComment{N $\times$ D} \\
        \PyCode{p'' = torch.reshape(p'', [N, D])} \PyComment{N $\times$ D} \\
        \PyCode{p = torch.einsum('np,nq->pq', [p', p'']) / N} \PyComment{compute empirical joint distribution} \\
        \PyCode{p\_s = select(p)} \\
        \PyCode{loss\_ent = (p\_s * torch.log(p\_s)).sum() / (S $\times$ S)} \\
        \text{ } \\
        \PyComment{final loss} \\
        \PyCode{loss = loss\_ent + lambda * loss\_TI} \PyComment{lambda=1 by default} \\
        \text{ } \\
        \PyComment{optimization step} \\
        \PyCode{loss.backward()} \\
        \PyCode{optimizer.step()} \\
    \Indm 
\caption{PyTorch-style pseudocode for MUSIC}
\label{algo:code}
\end{algorithm}

\section{Theoretical Analysis}

\textbf{The optimal solution to the first part in $L_{ent}$.}
As described in Section \ref{sec_analysis},
the entropy loss function consists of two parts: (1) the entropy over diagonal elements and (2) the entropy over the elements of off-diagonal blocks, as illustrated in Fig. \ref{fig_ssl}.
Now, let us minimize the first part, which is formulated as
\begin{equation}
    L_{ent} = \frac{1}{S}\sum_{s',s'', s'=s''}\sum_{d', d'', d'=d''} p(s', s'', d', d'') \log(p(s', s'', d', d''))
\end{equation}
Since every diagonal block has the same optimal solution, here we can only consider the $s^{th}$ diagonal block, and the object function can be simplified as
\begin{equation}
    L_{ent}(s, s) = \sum_{d=1}^{D_S} p(s, s, d, d) \log(p(s, s, d, d))
\end{equation}
where $0 \le p(s, s, d, d) \le 1$, $0 \le \sum_{d=1}^{D_S} p(s, s, d, d) \le 1$. Then, it is easy to find the solution that minimizes this objective function, i.e., $\forall s, d, p(s, s, d, d)=\frac{1}{D_S}$, or $\forall s', s'', s'=s'', d', d'', d'=d'', p(s', s'', d', d'')=\frac{1}{D_S}$.

As defined in Eqs. (\ref{eq_prob}) and (\ref{eq_joint}), we have $\forall s', s'', d', d''$, $0 \le p'_i(s', d') \le 1, 0 \le p''_i(s'', d'') \le 1, \sum_{d'=1}^{D_S} p'_i(s', d') = 1, \sum_{d''=1}^{D_S} p''_i(s'', d'') = 1$, and $ p(s', s'', d', d'') = \frac{1}{N}\sum_{i=1}^N p'_i(s', d') p''_i(s'', d'')$.

Thus, $\forall s', s'', d', d'', d'=d''$, we have

\begin{equation}
    \sum_{d',d'', d'=d''} p(s', s'', d', d'') = \sum_{d',d'', d'=d''} \frac{1}{D_S} = \sum_{d=1}^{D_S} \frac{1}{D_S} = 1
    \label{eq_a9}
\end{equation}

Given the above derived results, let us next prove that for $\forall s', s'', s'=s''$, $\exists d',d'', d'=d''$, $p'_i(s',d)=p''_i(s'', d)=1$ by contradiction.

If its negation is true, i.e., $\forall s', s'', s'=s''$, if $\exists d', d'', d'=d''$, either $0 < p'_i(s',d') < 1$ or $0< p''_i(s'', d'') < 1$, then we have either $p'_i(s',d') < \sum_{d'=1}^{D_S}p_i(s', d') =1$ or $p''_i(s'',d'') < \sum_{d''=1}^{D_S}p_i(s'', d'') =1$. When  $p'_i(s',d') < \sum_{d'=1}^{D_S}p_i(s', d') =1$, then we have
\begin{equation}
    \begin{split}
        \sum_{d',d'', d'=d''} p(s', s'', d', d'') &= \sum_{d',d'', d'=d''} \frac{1}{N}\sum_{i=1}^N p'_i(s', d') p''_i(s'', d'') \\
                                                  &= \frac{1}{N}\sum_{i=1}^N \sum_{d',d'', d'=d''}p'_i(s', d') p''_i(s'', d'') \\
                                                  &= \frac{1}{N}\sum_{i=1}^N \sum_{d}^{D_S}p'_i(s', d) p''_i(s'', d) \\
                                                  &< \frac{1}{N}\sum_{i=1}^N \sum_{d}^{D_S}p'_i(s', d) \sum_{d}^{D_S}p''_i(s'', d) \\
                                                  &=1
    \end{split}
    \label{eq_a10}
\end{equation}
That is, $\sum_{d',d'', d'=d''} p(s', s'', d', d'')<1$, which leads to a contradiction with Eq. (\ref{eq_a9}). Similarly, when $p''_i(s'',d'') < \sum_{d''=1}^{D_S}p_i(s'', d'') =1$, we have the same conclusion. Therefore, the statement that $\forall s', s'', s'=s''$, if $d'=d''$, then $p'_i(s',d')=p''_i(s'', d'')=1$ is true. It means that for $\forall s$, $p'_i(s, :)$ and $p''_i(s, :)$ are one-hot vectors and equals to each other.

Because $\forall s, d, p(s, s, d, d)=\frac{1}{D_S}, p'(s, d)=p''(S, d)$, and $p'(S, d)$ and $p''(S, d)$ are one-hot vectors, then $p(s, s, d, d)=\frac{1}{N}\sum_{i=1}^{N}p'_i(s,d)p''_i(s,d)=\frac{1}{N}\sum_{i=1}^{N}p'_i(s,d)=\frac{1}{D_S}$. This each segment evenly assigns samples into each unit.


\textbf{Covariance of the optimal solution.}
The optimal solution to maximize the joint entropy over the off-diagonal blocks for the second part is $p(s', s'', d', d'') = \frac{1}{(D_S)^2}, \forall s' \ne s''$. According to the above proof that each segment evenly assigns samples into each unit, then $\mathbbm{E}[p(s'', d'')]=\frac{1}{D_S}$. We can theoretically demonstrate the covariance between any two bins from different segments is zero. Specifically, $\forall s', s'', s'\ne s'', d', d''$, we have
\begin{equation}
\begin{split}
    \text{cov}[p(s', d'), p(s'', d'')] &= \mathbbm{E}[p(s', d') p(s'', d'')] - \mathbbm{E}[p(s', d')]\mathbbm{E}[p(s'', d'')] \\
                                     &= p(s', d', s'', d'') - \frac{1}{D_S}\times \frac{1}{D_S} \\
                                     &= \frac{1}{(D_S)^2} - \frac{1}{D_S}\times \frac{1}{D_S} = 0.
\end{split}
\end{equation}
Furthermore, any two units within the same segment are negatively correlated. Formally, $\forall s, d' \ne d''$, we have
\begin{equation}
    \begin{split}
        \text{cov}[p(s, d'), p(s, d'')] &= \mathbbm{E}[p(s, d') p(s, d'')] - \mathbbm{E}[p(s, d')]\mathbbm{E}[p(s, d'')] \\
                                         &= p(s, d', s, d'') - \frac{1}{D_S}\times \frac{1}{D_S} \\
                                         &= 0 - \frac{1}{D_S}\times \frac{1}{D_S} = -\frac{1}{{D_S}^2}.
    \end{split}
\end{equation}
That is, the unit within each segment encodes discriminative features, while the units from different segments encode unrelated and diverse features.



\bibliographystyle{unsrt}  
\bibliography{references}

\begin{thebibliography}{10}

\bibitem{bengio2013representation}
Yoshua Bengio, Aaron Courville, and Pascal Vincent.
\newblock Representation learning: A review and new perspectives.
\newblock {\em IEEE transactions on pattern analysis and machine intelligence},
  35(8):1798--1828, 2013.

\bibitem{hendrycks2019using}
Dan Hendrycks, Mantas Mazeika, Saurav Kadavath, and Dawn Song.
\newblock Using self-supervised learning can improve model robustness and
  uncertainty.
\newblock {\em Advances in Neural Information Processing Systems}, 32, 2019.

\bibitem{mohseni2020self}
Sina Mohseni, Mandar Pitale, JBS Yadawa, and Zhangyang Wang.
\newblock Self-supervised learning for generalizable out-of-distribution
  detection.
\newblock In {\em Proceedings of the AAAI Conference on Artificial
  Intelligence}, volume~34, pages 5216--5223, 2020.

\bibitem{hadsell2006dimensionality}
Raia Hadsell, Sumit Chopra, and Yann LeCun.
\newblock Dimensionality reduction by learning an invariant mapping.
\newblock In {\em 2006 IEEE Computer Society Conference on Computer Vision and
  Pattern Recognition (CVPR'06)}, volume~2, pages 1735--1742. IEEE, 2006.

\bibitem{van2018representation}
Aaron Van~den Oord, Yazhe Li, and Oriol Vinyals.
\newblock Representation learning with contrastive predictive coding.
\newblock {\em arXiv e-prints}, pages arXiv--1807, 2018.

\bibitem{simclr}
Ting Chen, Simon Kornblith, Mohammad Norouzi, and Geoffrey Hinton.
\newblock A simple framework for contrastive learning of visual
  representations.
\newblock In {\em Proceedings of the 37th International Conference on Machine
  Learning}, volume 119, pages 1597--1607, 2020.

\bibitem{moco}
Kaiming He, Haoqi Fan, Yuxin Wu, Saining Xie, and Ross Girshick.
\newblock Momentum contrast for unsupervised visual representation learning.
\newblock In {\em Proceedings of the IEEE/CVF Conference on Computer Vision and
  Pattern Recognition (CVPR)}, June 2020.

\bibitem{wu2018unsupervised}
Zhirong Wu, Yuanjun Xiong, Stella~X Yu, and Dahua Lin.
\newblock Unsupervised feature learning via non-parametric instance
  discrimination.
\newblock In {\em Proceedings of the IEEE conference on computer vision and
  pattern recognition}, pages 3733--3742, 2018.

\bibitem{misra2020self}
Ishan Misra and Laurens van~der Maaten.
\newblock Self-supervised learning of pretext-invariant representations.
\newblock In {\em Proceedings of the IEEE/CVF Conference on Computer Vision and
  Pattern Recognition}, pages 6707--6717, 2020.

\bibitem{grill2020bootstrap}
Jean-Bastien Grill, Florian Strub, Florent Altch{\'e}, Corentin Tallec, Pierre
  Richemond, Elena Buchatskaya, Carl Doersch, Bernardo Avila~Pires, Zhaohan
  Guo, Mohammad Gheshlaghi~Azar, et~al.
\newblock Bootstrap your own latent-a new approach to self-supervised learning.
\newblock {\em Advances in Neural Information Processing Systems},
  33:21271--21284, 2020.

\bibitem{Chen_2021_CVPR}
Xinlei Chen and Kaiming He.
\newblock Exploring simple siamese representation learning.
\newblock In {\em Proceedings of the IEEE/CVF Conference on Computer Vision and
  Pattern Recognition (CVPR)}, pages 15750--15758, June 2021.

\bibitem{understand}
Tongzhou Wang and Phillip Isola.
\newblock Understanding contrastive representation learning through alignment
  and uniformity on the hypersphere.
\newblock In {\em Proceedings of the 37th International Conference on Machine
  Learning}, volume 119, pages 9929--9939, 2020.

\bibitem{zhang2021does}
Chaoning Zhang, Kang Zhang, Chenshuang Zhang, Trung~X Pham, Chang~D Yoo, and
  In~So Kweon.
\newblock How does simsiam avoid collapse without negative samples? a unified
  understanding with self-supervised contrastive learning.
\newblock In {\em International Conference on Learning Representations}, 2021.

\bibitem{richemond2020byol}
Pierre~H Richemond, Jean-Bastien Grill, Florent Altch{\'e}, Corentin Tallec,
  Florian Strub, Andrew Brock, Samuel Smith, Soham De, Razvan Pascanu, Bilal
  Piot, et~al.
\newblock Byol works even without batch statistics.
\newblock {\em arXiv preprint arXiv:2010.10241}, 2020.

\bibitem{tian2021understanding}
Yuandong Tian, Xinlei Chen, and Surya Ganguli.
\newblock Understanding self-supervised learning dynamics without contrastive
  pairs.
\newblock In {\em International Conference on Machine Learning}, pages
  10268--10278. PMLR, 2021.

\bibitem{caron2018deep}
Mathilde Caron, Piotr Bojanowski, Armand Joulin, and Matthijs Douze.
\newblock Deep clustering for unsupervised learning of visual features.
\newblock In {\em Proceedings of the European conference on computer vision
  (ECCV)}, pages 132--149, 2018.

\bibitem{asano2019self}
Yuki~Markus Asano, Christian Rupprecht, and Andrea Vedaldi.
\newblock Self-labelling via simultaneous clustering and representation
  learning.
\newblock {\em arXiv preprint arXiv:1911.05371}, 2019.

\bibitem{caron2020unsupervised}
Mathilde Caron, Ishan Misra, Julien Mairal, Priya Goyal, Piotr Bojanowski, and
  Armand Joulin.
\newblock Unsupervised learning of visual features by contrasting cluster
  assignments.
\newblock {\em Advances in Neural Information Processing Systems},
  33:9912--9924, 2020.

\bibitem{ermolov2021whitening}
Aleksandr Ermolov, Aliaksandr Siarohin, Enver Sangineto, and Nicu Sebe.
\newblock Whitening for self-supervised representation learning.
\newblock In {\em International Conference on Machine Learning}, pages
  3015--3024. PMLR, 2021.

\bibitem{zbontar2021barlow}
Jure Zbontar, Li~Jing, Ishan Misra, Yann LeCun, and St{\'e}phane Deny.
\newblock Barlow twins: Self-supervised learning via redundancy reduction.
\newblock In {\em International Conference on Machine Learning}, pages
  12310--12320. PMLR, 2021.

\bibitem{bardes2021vicreg}
Adrien Bardes, Jean Ponce, and Yann LeCun.
\newblock Vicreg: Variance-invariance-covariance regularization for
  self-supervised learning.
\newblock {\em arXiv preprint arXiv:2105.04906}, 2021.

\bibitem{ji2019invariant}
Xu~Ji, Joao~F Henriques, and Andrea Vedaldi.
\newblock Invariant information clustering for unsupervised image
  classification and segmentation.
\newblock In {\em Proceedings of the IEEE/CVF International Conference on
  Computer Vision}, pages 9865--9874, 2019.

\bibitem{peng2005feature}
Hanchuan Peng, Fuhui Long, and Chris Ding.
\newblock Feature selection based on mutual information criteria of
  max-dependency, max-relevance, and min-redundancy.
\newblock {\em IEEE Transactions on pattern analysis and machine intelligence},
  27(8):1226--1238, 2005.

\bibitem{He_2016_CVPR}
Kaiming He, Xiangyu Zhang, Shaoqing Ren, and Jian Sun.
\newblock Deep residual learning for image recognition.
\newblock In {\em Proceedings of the IEEE Conference on Computer Vision and
  Pattern Recognition (CVPR)}, June 2016.

\bibitem{you2017large}
Yang You, Igor Gitman, and Boris Ginsburg.
\newblock Large batch training of convolutional networks.
\newblock {\em arXiv preprint arXiv:1708.03888}, 2017.

\bibitem{goyal2017accurate}
Priya Goyal, Piotr Doll{\'a}r, Ross Girshick, Pieter Noordhuis, Lukasz
  Wesolowski, Aapo Kyrola, Andrew Tulloch, Yangqing Jia, and Kaiming He.
\newblock Accurate, large minibatch sgd: Training imagenet in 1 hour.
\newblock {\em arXiv preprint arXiv:1706.02677}, 2017.

\bibitem{loshchilov2016sgdr}
Ilya Loshchilov and Frank Hutter.
\newblock Sgdr: Stochastic gradient descent with warm restarts.
\newblock {\em arXiv preprint arXiv:1608.03983}, 2016.

\end{thebibliography}

\end{document}